%% file: main.tex
\documentclass[conference]{IEEEtran}
\usepackage{graphicx} 
\graphicspath{ {./figs/} }
\usepackage{caption}
\usepackage{subcaption}
\usepackage{url}
\usepackage{subcaption}
\usepackage{multirow}
\usepackage{float}
\usepackage{tikz}
\usepackage{amsmath}
\usepackage{amssymb}
\usepackage{amsfonts}
\usepackage{soul}
\def\BibTeX{{\rm B\kern-.05em{\sc i\kern-.025em b}\kern-.08em
    T\kern-.1667em\lower.7ex\hbox{E}\kern-.125emX}}

\title{Improving Internet Traffic Matrix Prediction via Time Series Clustering}

\author{\IEEEauthorblockN{Martha Cash}
\IEEEauthorblockA{
\textit{Worcester Polytechnic Institute}\\
Worcester, MA\\
mcash@wpi.edu}
\and
\IEEEauthorblockN{Alexander Wyglinski}
\IEEEauthorblockA{
\textit{Worcester Polytechnic Institute}\\
Worcester, MA\\
alexw@wpi.edu}}

\begin{document}

\maketitle
\vspace{-5pt}

\begin{abstract}
We present a novel framework that leverages time series clustering to improve internet traffic matrix (TM) prediction using deep learning (DL) models. Traffic flows within a TM often exhibit diverse temporal behaviors, which can hinder prediction accuracy when training a single model across all flows. To address this, we propose two clustering strategies —source clustering and histogram clustering —that group flows with similar temporal patterns prior to model training. Clustering creates more homogeneous data subsets, enabling models to capture underlying patterns more effectively and generalize better than global prediction approaches that fit a single model to the entire TM. Compared to existing TM prediction methods, our method reduces RMSE by up to 92\% for Abilene and 75\% for G\'EANT. In routing scenarios, our clustered predictions also reduce maximum link utilization (MLU) bias by 18\% and 21\%, respectively, demonstrating the practical benefits of clustering when TMs are used for network optimization.
\end{abstract}

\section{Introduction}\label{sec:Intro}
\input{sections/introduction}

\section{Related Work}\label{sec:related_work}
\input{sections/related_work}

\section{Traffic Matrix Prediction Overview}\label{sec:overview}
\input{sections/overview}

\section{Proposed Clustering Approaches for Traffic Matrix Prediction}\label{sec:clustering}
\input{sections/clustering}

\section{Experimental Setup}\label{sec:experiments}
\input{sections/experiments}

\section{Results}\label{sec:results}
\input{sections/results}

\section{Conclusion}\label{sec:conclusion}
\input{sections/conclusion}

\section*{Acknowledgments}\label{sec:conclusion}
\input{sections/acknowledgments}

\end{document}

%% file: sections/introduction.tex
 A TM quantifies traffic demand between source-destination (SD) node pairs in a network, and are essential for network planning, serving as critical input for many traffic engineering (TE) tasks, such as routing optimization and network resource allocation~\cite {soule2005}. However, TMs are challenging to measure and collect in real time~\cite{soule2005}, motivating the need for accurate TM prediction to support TE tasks~\cite{azzouni2018neutm}.

Recent research trends in TM prediction focus on deep learning (DL) models~\cite{azzouni2018neutm, aloraifan2021deep, jiang2022internet}, which exploit the ability of DL to model complex nonlinear relationships in traffic data. These models demonstrate strong performance, as indicated by error metrics such as Root Mean Squared Error (RMSE) and Mean Absolute Error (MAE). Most approaches adopt a global forecasting paradigm, training a single model to predict all traffic demands in a TM simultaneously. As shown by~\cite{MonteroManso2020}, global models can outperform local models—which predict each traffic demand independently—when the dataset features exhibit similar patterns. However,~\cite{Bandara2017} argues that heterogeneous features can lead to decreased accuracy in global models.

\begin{figure}[!t]
    \centering
    \includegraphics[width=0.48\textwidth]{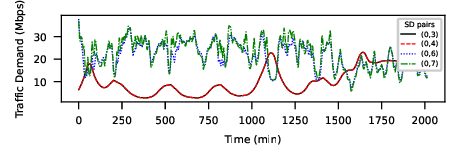}
    \caption{Traffic flows from different source destination pairs in the Abilene~\cite{AbileneDataset} dataset displaying varying behavior.}
    \setlength{\belowcaptionskip}{-30pt}
    \label{fig:traffic_flows}
    \vspace{-15pt}
\end{figure}

Referring to Fig.~\ref{fig:traffic_flows}, we observe that traffic demands within a TM often differ significantly in behavior. Some demands, SD pairs $(0,3)~\&~(0,4)$, exhibit similar trends, while others, SD pairs $(0,4)~\&~(0,7)$, deviate markedly. The heterogeneity among traffic flows has led to the development of complex DL architectures, such as Attention-based Convolutional Recurrent Neural Networks (ARCNN)~\cite{gao2020incorporating} and Convolutional Long Short-Term Memory Networks (CNN-LSTM)~\cite{Jiang2022}, designed to capture the diverse flow dynamics in TMs.

In this work, we challenge the growing trend of designing complex global models for TM prediction. Instead, we propose exploiting the inherent similarities among subsets of traffic flows through clustering. Clustering is an unsupervised learning technique that partitions time series into groups with high intra-group similarity and low inter-group similarity~\cite{Paparrizos2024}. Rather than fitting a single model to predict the entire TM, we train separate models for each cluster of flows.

The main contributions of this paper are:
\begin{itemize}
\item Two novel clustering methods for TMs: source-based clustering and histogram-based clustering.
\item A comparison of the clustering-based models against three state-of-the-art (SOTA) TM prediction baselines.
\item TM prediction performance evaluation of both clustering approaches with quantitative error metrics (RMSE and MAE) and in an applied routing context.
\end{itemize}

The remainder of the paper is organized as follows. Section~\ref{sec:related_work} reviews related work in TM prediction and time series clustering. Section~\ref{sec:overview} overviews TMs and TM prediction. Section~\ref{sec:clustering} details our proposed clustering approaches. Section~\ref{sec:experiments} outlines the datasets and experimental setup. Section~\ref{sec:results} presents the evaluation results. Finally, Section~\ref{sec:conclusion} concludes the paper and outlines directions for future work.

%% file: sections/related_work.tex
Internet traffic prediction is a time series forecasting problem. A wide range of models, spanning from classical statistical methods to DL approaches, have been employed in this context. Otoshi \textit{et al.} decompose traffic into long-term and short-term components and apply the seasonal autoregressive integrated moving average model for prediction~\cite{otoshi2015traffic}. However, this method requires repeatedly differencing the data to achieve stationarity, which limits its practical use. Valadarsky \textit{et al.} explore Feedforward Neural Networks, Convolutional Neural Networks (CNNs), and nonlinear autoregressive models for traffic prediction, but these fail to capture temporal dependencies, leading to poor predictive performance for TM~\cite{valadarsky2017learning}.

To better model temporal dynamics, recent work~\cite{zhao2018towards, troia2018deep, ramakrishnan2018network} has turned to Recurrent Neural Networks (RNNs), such as Long Short-Term Memory (LSTM) and Gated Recurrent Unit (GRU) models, to predict entire TMs. Other approaches, such as~\cite{le2021ai, zheng2022flow}, treat each TM element as an independent time series and train separate models for each. Liu \textit{et al.} propose a hybrid approach that predicts total traffic volume and distributes it across TM elements using precomputed ratios, while correcting for high-volume flows with specialized models~\cite{liu2014prediction}. However, several studies~\cite{liu2019traffic, gao2020incorporating} have demonstrated that modeling TM elements independently overlooks their correlations, which can potentially compromise overall prediction performance.

TMs can be interpreted as a structured collection of interrelated time series. In this broader context, several works have explored clustered time series forecasting. Bandara \textit{et al.} address the challenge of forecasting across heterogeneous time series by first clustering series based on statistical properties (e.g., mean, variance, autocorrelation), and then fitting RNN models per cluster~\cite{Bandara2017}. Martinez \textit{et al.} use geometric distances to cluster time series and apply $k$-nearest neighbors for forecasting~\cite{Martinez2019}. Zhang \textit{et al.} propose clustering traffic flows in transportation networks using flow probability distributions, and Zou \textit{et al.} apply hierarchical clustering and CNNs to predict lane traffic volumes in transportation systems~\cite{Zhang2022, Zou2023}.

No prior work has applied time series clustering to internet TMs. This paper aims to bridge that gap by clustering similar traffic flows in TMs and training dedicated models for each cluster, thus exploring the potential to improve prediction accuracy while reducing model complexity.

%% file: sections/overview.tex
For a network with $N$ nodes, a TM is an $N \times N$ matrix where each entry $(i,j)$ represents the traffic volume from source node $i$ to destination node $j$, typically in bytes. At time interval $\Delta t$, the TM is denoted $\mathrm{TM}_t$, where $t \in [t, t + \Delta t] \subset T$ and $T$ is the full measurement period. A TM at time $t$ is:

\begin{equation}
    \label{eq:traffic_matrix}
    \mathrm{TM_t} = \begin{bmatrix}
    T_{1,1}^t & T_{1,2}^t & \cdots & T_{1,n}^t \\
    T_{2,1}^t & T_{2,2}^t & \cdots & T_{2,n}^t \\
    \vdots    & \vdots    & \ddots & \vdots    \\
    T_{n,1}^t & T_{n,2}^t & \cdots & T_{n,n}^t \\
    \end{bmatrix}
\end{equation}
Over the full interval $T$, each TM entry $\mathrm{TM}[i,j]$ becomes a time series of length $T$ capturing traffic from node $i$ to $j$, referred to as a traffic flow, and each $(i,j)$ pair is called a source-destination (SD) pair.

The goal of TM prediction is to forecast $\mathrm{TM}_{t+1}$ given a history of $k$ previous matrices: $(\mathrm{TM}_{t-k}, \dots, \mathrm{TM}_t)$. TMs exhibit nonlinear, nonstationary, and periodic behavior, making prediction challenging. DL models are well-suited for this task due to their ability to capture such complex patterns. We formulate prediction as a supervised learning problem, following the approach in~\cite{valadarsky2017learning}.

To construct training data, we apply a sliding window over $D$ consecutive TMs to generate $W = D - L + 1$ windows of length $L$~\cite{azzouni2018neutm}. In each window, the first $L-1$ matrices serve as inputs and the $L^{th}$ matrix is the prediction target. Repeating this across the TMs yields a training set of input-output pairs for the DL model.

TM prediction approaches fall into two categories: entire-matrix (EM) prediction and local prediction~\cite{liu2014prediction, zheng2022flow, azzouni2018neutm}. EM prediction uses a single model to forecast the whole TM, while local prediction trains one model per flow, resulting in $N^2$ models for a network with $N$ nodes. Although local prediction often yields higher accuracy~\cite{zheng2022flow}, it is computationally expensive and scales poorly. EM models can match local performance~\cite{liu2014prediction}, but usually require complex, resource-intensive architectures.

A key limitation of EM prediction is its implicit assumption of homogeneity among flows, an assumption that rarely holds in practice. Gao \textit{et al}.~\cite{gao2020incorporating} show that flows from the same source are often correlated, while flows from different sources may not be. This indicates that TMs contain latent structure that can be exploited for prediction.

%% file: sections/clustering.tex
We propose a third prediction approach for TM prediction: cluster-based prediction using time series clustering. Instead of modeling each flow or the entire matrix, we group similar flows and train one model per cluster to reduce the model count compared to local prediction while preserving flow-specific dynamics better than EM models. In this work, we study two clustering methods: source-based, which groups flows by their origin node, and histogram-based, which clusters flows by their distributional similarity.

\subsection{Source-Based Clustering}
\begin{figure}[!t]
\centering
\begin{subfigure}[t]{0.479\columnwidth}
  \includegraphics[width=\linewidth]{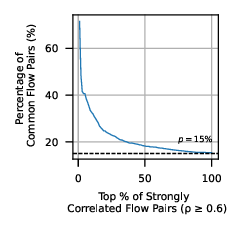}
  \caption{Abilene}
\end{subfigure}\hfill 
\begin{subfigure}[t]{0.479\columnwidth}
  \includegraphics[width=\linewidth]{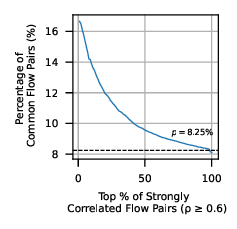}
  \caption{G\'EANT}
\end{subfigure}
\caption{The proportion of flow pairs with the same source in the top x\% strongly-correlated flow pairs. Strongly correlated flows decreases as more pairs are included. The dashed line indicates the baseline percentage. Abilene has a baselines percentage of 15\% traffic flow pairs being strongly correlated. G\'EANT has a baseline percentage of 8.25\%}
\label{fig:corr}
\vspace{-15pt}
\end{figure}

The first clustering approach, source-based clustering, is motivated by \cite{gao2020incorporating}. Traffic flows from the same source node should exhibit similar characteristics, such as periodicity, because the traffic demands are generated by the same sources~\cite{TMPrimer}. To demonstrate this, we compute the correlation ($\rho$) among flows. The correlation reflects the relationship between two random variables, and the Pearson Correlation Coefficient is a well known way of measuring the correlation~\cite{taylor1990correlation} . The correlation among flows is calculated following: 

\begin{equation}
 \rho = \frac{cov(\mathbf{TM}[i,j], \mathbf{TM}[i^{'}, j^{'}])}{\sigma_{\mathbf{TM}[i,j]}\sigma_{\mathbf{TM}[i^{'}, j^{'}]}}
\end{equation}
Where $cov(\cdot)$ is the covariance function and $\sigma_{\mathbf{TM}[i,j]}$ is the variance of the flow $\mathbf{TM}[i,j]$ across the entire measurement interval $T$. The variable $\rho$ varies between $[-1, 1]$. The closer to 1, the more positively correlated the two flows are, and the closer to 0, the more negatively correlated they are. Fig.~\ref{fig:corr} demonstrates that the traffic flows are strongly correlated ($\rho \geq 0.6$) when they come from the same source using two real-world datasets, Abilene and G\'EANT. Further details of the datasets are provided in Section~\ref{sec:experiments}. 

Fig.~\ref{fig:corr} shows the temporal consistency of strongly correlated flow pairs ($\rho \geq 0.6$) for Abilene and G\'EANT. In Abilene, a significant fraction of flow pairs remain highly correlated over time, especially the top 15\%, indicating a stable core of traffic relationships. G\'EANT, by contrast, shows lower consistency and a smaller stable core, suggesting more transient and variable flow behavior. This contrast highlights the need to model persistent dependencies in stable networks, such as Abilene, while remaining adaptable in dynamic ones, like G\'EANT. The concentration of stable correlations among a small subset of flows, particularly in Abilene, suggests dominant sources often drive traffic. This supports source-based clustering, which groups flows by origin to exploit shared temporal patterns and improve prediction robustness where stable correlations exist.

\subsection{Histogram-Based Clustering}
\begin{figure}[!t]
\centering
    \begin{subfigure}[t]{0.479\columnwidth}
        \includegraphics{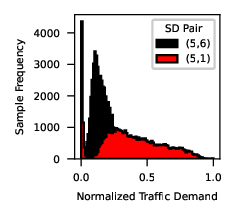}
        \caption{Abilene}
    \end{subfigure}
    \hfill
    \begin{subfigure}[t]{0.479\columnwidth}
        \includegraphics{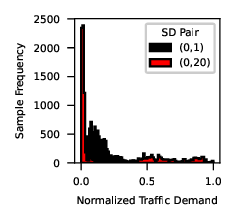}
        \caption{G\'EANT}
    \end{subfigure}
    \caption{Histograms of normalized traffic demand for two source-destination flows sharing a common source in (a) Abilene and (b) G\'EANT. The distributions show that flows originating from the same source can exhibit significantly different statistical behavior.}
    \label{fig:histograms}
    \vspace{-10pt}
\end{figure}

While source-based clustering leverages the intuition that flows originating from the same source node may share temporal patterns, it does not capture differences in traffic magnitude or variability. As shown in Fig.~\ref{fig:histograms}, flows such as $(0,6)$, $(0,7)$, $(0,3)$, and $(0,4)$ from the same source can exhibit markedly different traffic distributions, indicating that structural proximity does not always imply behavioral similarity.

To address this, we introduce a complementary histogram-based clustering approach that groups flows based on the distributional properties of their normalized traffic time series. By representing each flow as a histogram, we capture broader statistical characteristics, such as burstiness, flatness, or multi-modality, independent of source or destination. This enables grouping of flows with similar usage patterns, even across different nodes.

We quantify histogram similarity using the Jensen-Shannon divergence (JSD), a symmetric and bounded version of the Kullback-Leibler (KL) divergence. Given two flow histograms $P$ and $Q$, the JSD is defined as:

\begin{equation}\label{eq:JSD}
    JSD(P \lvert \rvert Q) = \frac{1}{2}D(P \lvert \rvert M) + \frac{1}{2}D(Q \lvert \rvert M)
\end{equation}

\noindent where $D(\cdot)$ denotes the KL divergence. JSD evaluates to 0 when $P$ and $Q$ are identical and reaches 1 when they are completely disjoint.

We compute pairwise JSD values between all flows, resulting in an $N^2 \times N^2$ distance matrix. Using hierarchical clustering, each flow initially forms its own cluster, and those with the smallest JSD are iteratively merged. This process produces a linkage matrix that captures the hierarchy of merges.

\begin{figure}[!h]
    \centering
    \begin{subfigure}{\columnwidth}
        \includegraphics{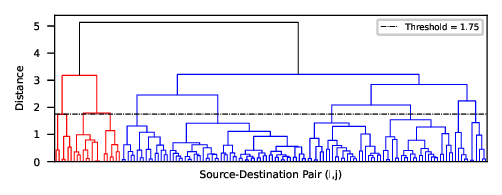}
        \caption{Abilene}
    \end{subfigure}
    \hfill
    \begin{subfigure}{\columnwidth}
        \includegraphics{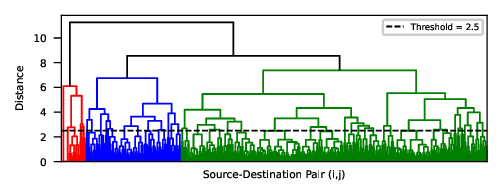}
        \caption{G\'EANT}
    \end{subfigure}
    \caption{Dendrograms for the Abilene and G\'EANT dataset visualizing the hierarchical clustering of the Jensen-Shannon divergence distance matrix. The dashed line represents the threshold for determining the number of clusters formed.}
    \label{fig:dendrogram}
    \vspace{-12pt}
\end{figure}

To select the number of clusters, we analyze the dendrograms in Fig.~\ref{fig:dendrogram} and cut at prominent vertical gaps that indicate natural divisions. We choose cut distances of 1.75 for Abilene and 2.5 for G\'EANT, yielding 9 and 47 clusters, respectively, thus balancing fragmentation and cluster granularity.

%% file: sections/experiments.tex
This section describes the datasets, preprocessing, benchmarks, and evaluation metrics used in our experiments.
\subsection{Datasets \& Preprocessing}
We evaluate our prediction methods using real data from the Abilene~\cite{AbileneDataset} and G\'{E}ANT ~\cite{GEANTDataset} networks. The Abilene network has 12 nodes and 30 links with TMs collected at 5-minute intervals over 24 weeks. The G\'{E}ANT network has 23 nodes and 74 links, with TMs collected on 15 minute intervals across 4 months. We apply an 80/20 training-testing split to the datasets, and 10\% of the training samples are used as a validation set. We use 10 TMs as historical input for prediction. We use min-max normalization to scale traffic flows to the range $[0, 1]$ for training.

\subsection{Benchmark Models}
We benchmark our clustering approaches against three SOTA models representing the unclustered baseline: Prophet~\cite{zhang2023prophet}, ARCNN~\cite{gao2020incorporating}, and a standalone GRU~\cite{azzouni2018neutm}. Prophet is a traffic engineering (TE)-centric framework that uses a GRU model with an angle-centric loss function for TM prediction. ARCNN is an SOTA spatial-temporal model that combines CNNs to capture inter-flow correlations, RNNs for intra-flow dependencies~\cite{ramakrishnan2018network}, and an attention mechanism for long-range temporal modeling. GRU, a type of RNN, is widely used in TM prediction for its ability to retain long-term dependencies. All three models operate under the EM prediction perspective. We also compare to the local prediction perspective and train $N^2$ GRU models (i.e., one per traffic flow)~\cite{zheng2022flow}. 

\subsection{Performance Metrics}
We evaluate the performance of the clustering approaches and the benchmarks using RMSE and MAE error metrics, defined as: 

\begin{equation}\label{eq:rmse}
    RMSE = \sqrt{\frac{1}{n}\sum_{i=1}^{n}(TM_{i}-\hat{TM}_i)^2}
\end{equation}

\begin{equation}\label{eq:MAE}
    MAE = \frac{1}{n}\sum_{i=1}^{n} |y_i-\hat{y}_i|
\end{equation}                                              
Where $n$ denotes the number of samples in the test set, $TM_i$ denotes the true TM, and $\hat{TM}_i$ denotes the predicted TM. 

We also assess predicted TMs in the context of traffic engineering (TE), which reduces congestion by routing flows to minimize maximum link utilization (MLU). TE is formulated as a multi-commodity flow problem using the TM as input \cite{valadarsky2017learning, zhang2023prophet, liu2019traffic}. Predicted TMs should yield MLU values close to those of the ground truth to be useful. Since small changes in flow ratios can cause significant MLU shifts, even low RMSE or MAE values may not accurately reflect TE performance. To evaluate this, we compute MLU bias and average MLU bias as follows: 

\begin{equation}\label{eq:bias}
      Bias = \frac{\hat{U}}{U}
\end{equation}

\begin{equation}\label{eq:avg_bias}
    \overline{Bias} = \frac{1}{n}\sum_{i=1}^{n} \frac{\hat{U}_i}{U_{i}}
\end{equation}
Where $\hat{U}$ is the MLU output from the predicted TM and $U$ is the MLU output from the ground truth TM. A bias closer to 1 indicates better prediction performance. We use the corresponding network topologies for Abilene and G\`EANT when computing the MLU for the predicted TMs.

\subsection{Implementation Details}
All DL models are implemented in PyTorch, and TE simulations use Gurobi. We use a one-layer GRU with a hidden size of 30 for all clustering approaches and benchmark models (excluding ARCNN), following the configurations used in prior work \cite{liu2019traffic}. Models are trained for 100 epochs with a batch size of 32, using the ADAM optimizer and a learning rate of 0.001. We use early stopping with a patience of 5 and a minimum delta of $1\text{e}{-5}$ to prevent overfitting. All parameter choices are based on settings commonly adopted in existing prediction approaches~\cite{valadarsky2017learning} - \cite{ramakrishnan2018network}

%% file: sections/results.tex
\begin{table*}[!t]
\caption{Normalized Average Test Error for TM Prediction Methods on Abilene and G\'EANT Networks}
\begin{center}
\begin{tabular}{|c|c|c|c|c|c|c|c|}
\hline
\textbf{Network} & \textbf{Metric} & \textbf{Histogram Clustering} & \textbf{Source Clustering} & \textbf{Prophet} & \textbf{ARCNN} & \textbf{GRU EM} & \textbf{GRU Local} \\
\hline
\multirow{2}{*}{Abilene} & RMSE & 2.38e-2 & 0.286 & 0.249 & 0.303 & 0.294 & 2.89e-3 \\
& MAE & 1.80e-2 & 0.242 & 0.198 & 0.253 & 0.249 & 1.88e-3 \\
\hline
\multirow{2}{*}{G\'EANT} & RMSE & 8.55e-2 & 0.114 & 0.354 & 0.257 & 0.244 & 1.11e-2 \\
& MAE & 6.79e-2 & 0.092 & 0.306 & 0.207 & 0.196 & 7.79e-3 \\
\hline
\end{tabular}
\label{tab:combined-error}
\end{center}
\end{table*}

Table~\ref{tab:combined-error} summarizes RMSE and MAE across prediction methods. Histogram-based clustering achieves the lowest error among EM models on both datasets, reducing RMSE by 92\% over ARCNN on Abilene and by 75\% over Prophet on G\'EANT. This highlights the benefit of grouping flows by distributional similarity, which yields more homogeneous training data and improves temporal modeling. While not outperforming local prediction, it offers substantial gains over EM models and remains far more scalable than training $N^2$ individual models. Source clustering provides modest improvements on Abilene but reduces RMSE by 67\% over Prophet on G\'EANT, indicating its potential on specific network topologies.

\begin{figure}
    \begin{subfigure}{\columnwidth}
        \includegraphics{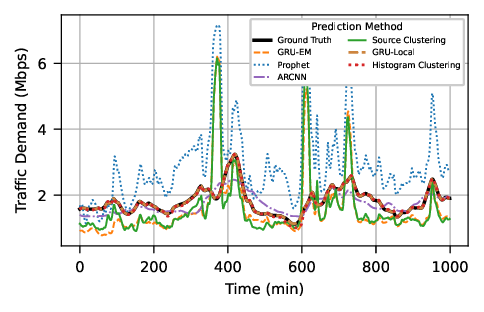}
        \caption{Predictions for Abilene traffic flow pair (6,3).}
        \vspace{-2pt}
    \end{subfigure}
    \hfill 
    \begin{subfigure}{\columnwidth}
        \includegraphics{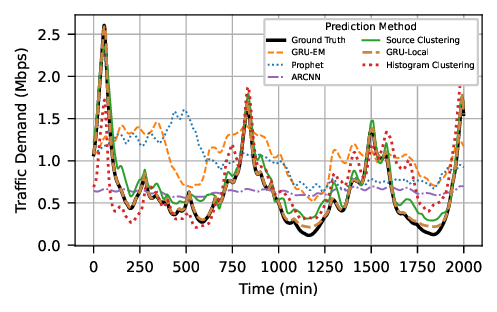}
        \caption{Predictions for G\'EANT traffic flow pair (6,16).}
    \end{subfigure}
    \caption{Traffic flows sampled from source destination pairs for different prediction methods.}
    \label{fig:results_preds}
    \vspace{-15pt}
\end{figure}

Fig.~\ref{fig:results_preds} highlights the benefits of clustering at the traffic flow level. For a sampled flow, histogram-based clustering (dotted red) closely tracks the ground truth (solid black), while EM methods often diverge due to bias toward dominant flows. On Abilene, source clustering (solid green) performs similarly to EM methods, suggesting the flow was grouped with high-volume flows. On G\'EANT, however, it more closely follows the ground truth and occasionally outperforms histogram clustering. This indicates that clustering effectiveness varies by topology and flow behavior; however, histogram clustering generally outperforms EM methods and matches local GRU models, demonstrating its advantages in terms of generalization and efficiency.

\begin{figure*}[!tb]
\centering
    \begin{subfigure}{\textwidth}
        \includegraphics{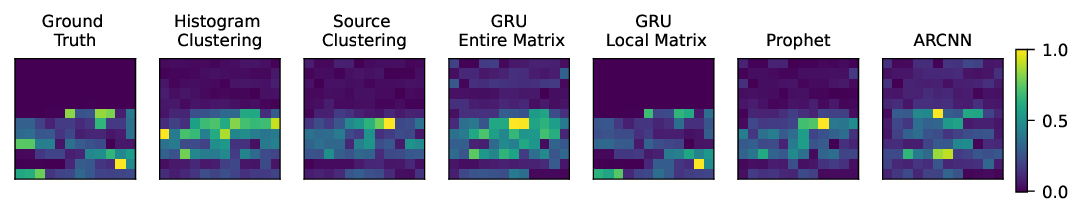}
        \caption{Abilene}
        \label{fig:Abilene_tm_pred}
    \end{subfigure}
    \hfill 
    \begin{subfigure}{\textwidth}
        \includegraphics{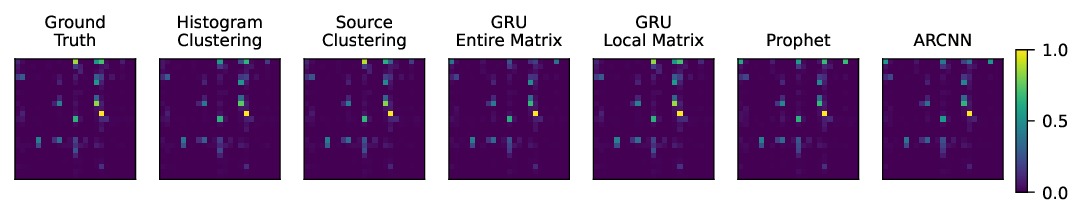}
        \caption{G\'EANT}
        \label{fig:Geant_tm_pred}
    \end{subfigure}
    \caption{Predicted traffic matrices compared to ground truth for two representative time steps.}
    \label{fig:results_preds_tm}

\end{figure*}

Fig.~\ref{fig:results_preds_tm} shows whole TM predictions, illustrating improved reconstruction quality over EM baselines. Histogram clustering better preserves spatial patterns under high load, as shown by the bright squares in Fig.~\ref{fig:Abilene_tm_pred}, capturing both the intensity and layout of significant flows. Source clustering performs similarly but slightly smooths localized hotspots. In sparse settings, shown in Fig.~\ref{fig:Geant_tm_pred}, both clustering methods outperform EM by retaining the location of low-volume flows and avoiding over-smoothing seen in GRU EM and Prophet. These results highlight the benefit of clustering for modeling traffic heterogeneity in both time and space.

Table~\ref{tab:mlu-bias} shows the impact of each method on routing performance. Histogram clustering yields an MLU bias of approximately 1 for both datasets, matching local prediction and outperforming the ARCNN model by 18\% on Abilene and by 21\% on G\'EANT, compared to the Prophet model or the GRU EM model. This indicates that it preserves key spatial and temporal flow characteristics. While source clustering performs worse on Abilene, it achieves strong results on G\'EANT, a larger and more complex network. 

\begin{table*}[!tb]
\caption{Average MLU Bias for Predicted Traffic Matrices On Abilene and G\'EANT Networks}
\begin{center}
\begin{tabular}{|c|c|c|c|c|c|c|}
\hline
\textbf{Network} & \textbf{Histogram Clustering} & \textbf{Source Clustering} & \textbf{Prophet} & \textbf{ARCNN} & \textbf{GRU EM} & \textbf{GRU Local}\\
\hline
Abilene & 1.00 & 0.80 & 0.92 & 0.82 & 0.92 & 1.02 \\
\hline
G\'EANT & 1.07 & 1.06 & 0.88 & 0.89 & 0.88 & 1.06 \\
\hline
\end{tabular}
\label{tab:mlu-bias}
\end{center}
\end{table*}

%% file: sections/conclusion.tex
We propose a clustering-based framework for Internet TM prediction, introducing two strategies: source clustering, which groups flows by source node, and histogram clustering, which groups flows by distributional similarity. We evaluate both on the Abilene and G\'EANT datasets, comparing them to three EM models and a local baseline that fits one model per flow.

Histogram clustering consistently outperforms EM models in terms of RMSE, MAE, and MLU bias, as it enables models to learn from more homogeneous flow groups, thereby improving generalization. While clustering does not fully match the accuracy of local prediction, it achieves comparable routing performance. It offers a better trade-off between accuracy and scalability, avoiding the cost of training $N^2$ models.

Future work includes exploring alternative distance metrics, fuzzy clustering methods, and strategies for selecting the optimal number of clusters.

%% file: sections/acknowledgments.tex
\noindent The authors would like to thank Prof. Randy Paffenroth for insightful conversations and valuable feedback that helped shape the development of this work.